\pdfoutput=1

\documentclass[11pt]{article}

\usepackage[]{emnlp2022}

\usepackage{times}
\usepackage{array}
\usepackage{latexsym}
\usepackage{microtype}
\usepackage{hyperref}

\usepackage{graphicx}
\usepackage{xspace}
\usepackage{paralist}
\usepackage{booktabs}
\usepackage{multirow}
\usepackage{amsmath}
\usepackage{amsfonts}
\usepackage{amssymb}
\usepackage{pifont}
\usepackage{mathrsfs}
\usepackage{algorithm,algpseudocode}
\usepackage{mdwlist}
\usepackage{enumitem}
\usepackage{color}
\usepackage{dsfont}
\usepackage{verbatim}
\usepackage{amsthm}
\usepackage{bm}
\usepackage{float}
\usepackage{caption}
\usepackage{subfigure}
\usepackage[T1]{fontenc}

\usepackage[utf8]{inputenc}

\usepackage{microtype}

\newcommand{\notes}[1]{}

\theoremstyle{definition}

\theoremstyle{plain}

\newcommand{\ith}[1]{\ensuremath{i^{{th}}}}

\newcount\permx
\newcount\permy
\def\permdot#1#2{
\permx=#1 \advance\permx by-1
\permy=#2 \advance\permy by-1
\psframe[fillcolor=black, fillstyle=solid]
(\permx,\permy)(#1, #2)
}

\newcommand{\boxnum}[1]{{\setlength{\fboxsep}{1pt}\raisebox{1pt}{\hspace{1pt}\fbox{\tiny #1}\hspace{1pt}}}}
\newcommand{\ind}[1]{\ensuremath{_{\kern-0.5pt\boxnum{#1}}}}

\newcommand{\smallnt}[1]{\ensuremath{_{\mbox{\tiny PP}}}\xspace}

\newcommand{\pseudocode}{Algorithm}
\floatname{algorithm}{\pseudocode}

\iffalse

\else

\fi

\title{Low-resource Neural Machine Translation with Cross-modal Alignment}

\author{
    Zhe Yang\textsuperscript{\rm 1,2},
    Qingkai Fang\textsuperscript{\rm 1,2},
    Yang Feng\textsuperscript{\rm 1,2}\thanks{ $\;\;$Corresponding author: Yang Feng.} \\
    \textsuperscript{\rm1} Key Laboratory of Intelligent Information Processing \\ Institute of Computing Technology, Chinese Academy of Sciences (ICT/CAS) \\
    \textsuperscript{\rm2} University of Chinese Academy of Sciences, Beijing, China \\
    \texttt{\{yangzhe22s1,fangqingkai21b,fengyang\}@ict.ac.cn} \\
}

\usepackage{lipsum}

\begin{document}
\maketitle

\begin{abstract}
How to achieve neural machine translation with limited parallel data? Existing techniques often rely on large-scale monolingual corpora, which is impractical for some low-resource languages. In this paper, we turn to connect several low-resource languages to a particular high-resource one by additional visual modality. Specifically, we propose a cross-modal contrastive learning method to learn a shared space for all languages, where both a coarse-grained sentence-level objective and a fine-grained token-level one are introduced. Experimental results and further analysis show that our method can effectively learn the cross-modal and cross-lingual alignment with a small amount of image-text pairs and achieves significant improvements over the text-only baseline under both zero-shot and few-shot scenarios. Our code could be found at \url{https://github.com/ictnlp/LNMT-CA}.


\end{abstract}

\section{Introduction}
Neural machine translation (NMT) has shown excellent performance and becomes the dominant paradigm of machine translation. However, NMT is a data-driven approach, which requires a large amount of parallel data. When the data is insufficient, it is impractical to train a reasonable NMT model. Unfortunately, there are many languages in the world for which sufficient training data is not available, and sometimes there is no parallel data at all. Therefore, the translation of low-resource languages is a vital challenge for NMT.

In recent years, researchers have attempted to improve the performance of NMT for low-resource languages. \citet{dblp1} proposed an unsupervised approach to learn weak mappings between languages with large amount of monolingual data (>1M), which is also costly for low-resource languages. \citet{DBLP:journals/tacl/LiuGGLEGLZ20, DBLP:conf/emnlp/LinPWQFZL20, pan-etal-2021-contrastive} proposed multilingual NMT models, which learn a shared space of multiple languages to achieve translations between languages that appear in the training set but do not have the corresponding parallel data. However, they still require auxiliary parallel data of source and target languages along with many other languages, which is still infeasible for low-resource languages.

\begin{figure}[tb]
    \centering
    \includegraphics[width=\linewidth]{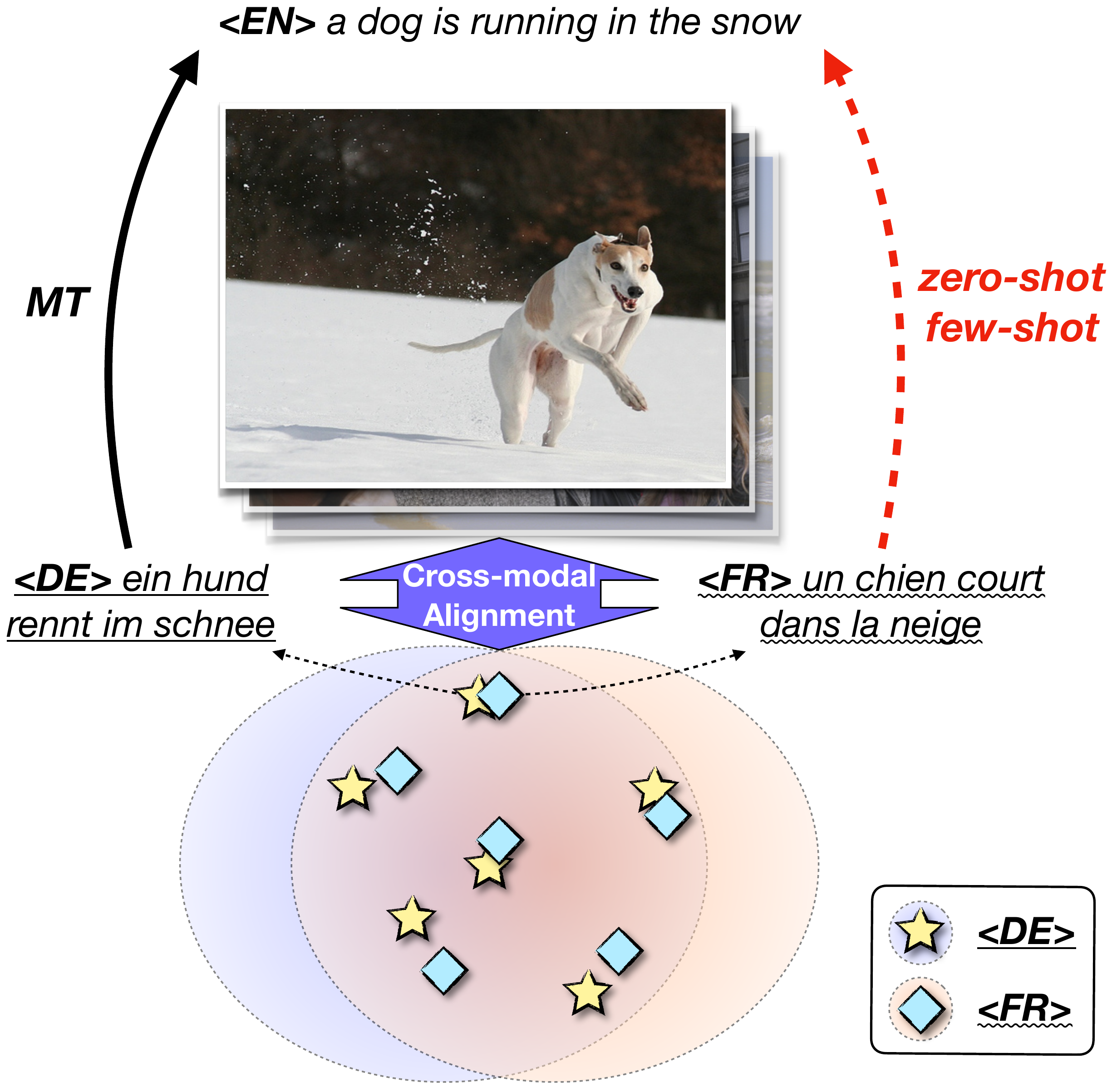}
    \caption{We aim at realizing zero-shot and few-shot machine translation for the low-resource language. Different languages with the same meanings are projected to a shared space by cross-modal alignment.}
    \label{fig:introduction}
\end{figure}

In recent years, with increasing attention of multi-modal tasks, resource of image-text pairs have become more abundant. Inspired by recent efforts on cross-modal alignment \citep{clip, DBLP:conf/acl/LiGNXLL0020, fang-etal-2022-STEMM}, in this paper, we propose a cross-modal contrastive learning method, which align different languages with images as the pivot to enable zero-shot and few-shot translations for low-resource languages. 
With parallel sentence pairs between one high-resource auxiliary language and the target language, we can achieve the translation from low-resource languages to the target language only by obtaining small amounts of image-text pairs (<0.1M) for those languages. 
The parallel sentence pairs are used to learn the mapping from the high-resource language to the target language, and the image-text pairs are used to learn a shared space for all languages through cross-modal alignment.
With images as the pivot, the mapping from the low-resource languages to the target language are learned, thus achieving zero-shot translation without any parallel sentence pairs between them.
As shown in Figure \ref{fig:introduction}, the high-resource language German and the low-resource language French are brought together by cross-modal alignment, which transfers the translation ability from DE$\rightarrow$EN to FR$\rightarrow$EN.
Experiments and analysis show that our method consistently outperforms the baseline under both zero-shot and few-shot scenarios. Furthermore, our method can effectively realize cross-modal and cross-lingual alignment.

\section{Method}

In this section, we present our proposed cross-modal contrastive learning method, which includes both sentence-level and token-level objectives.

\begin{figure*}[tb]
    \centering
    \includegraphics[width=\textwidth]{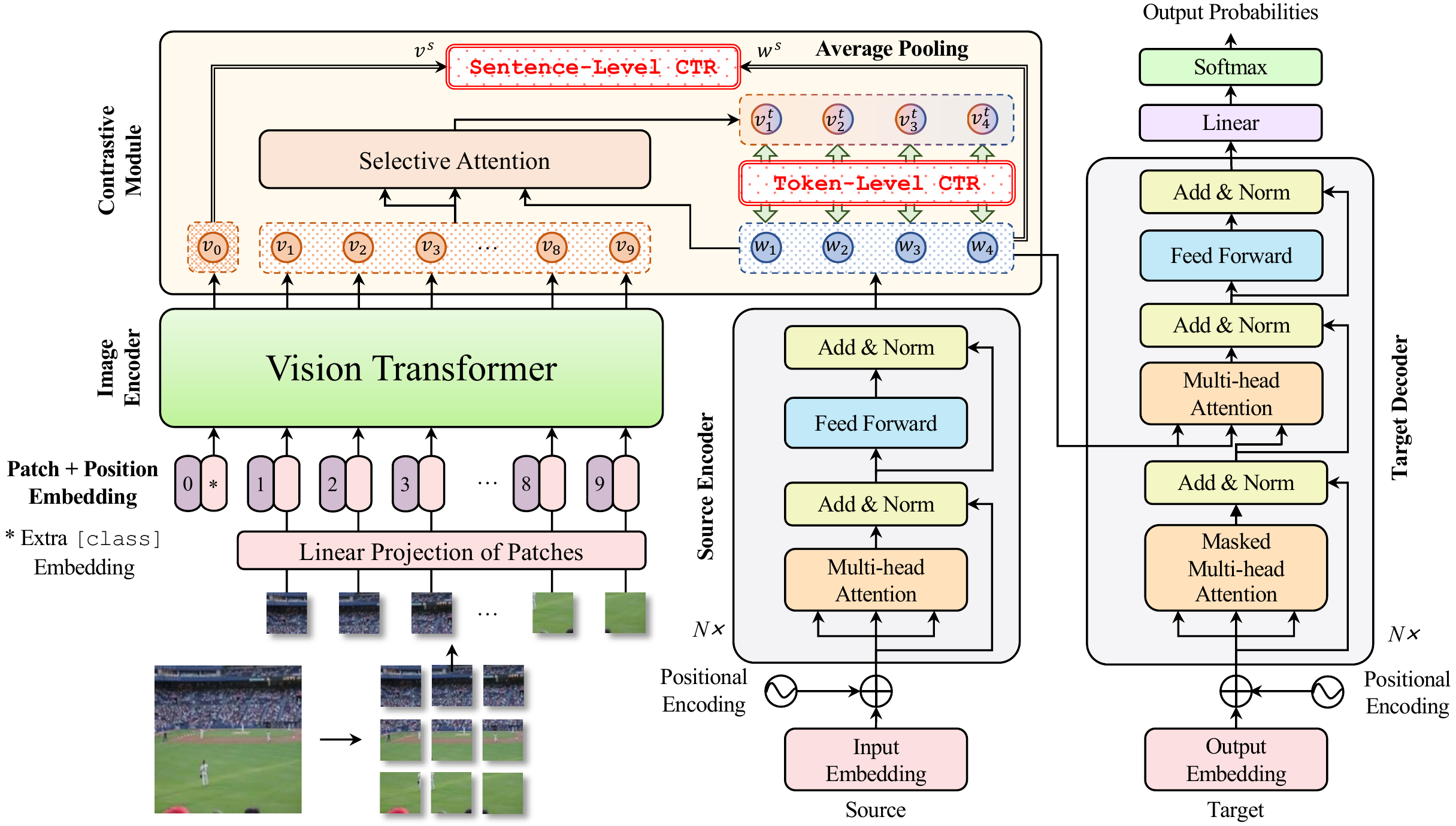}
    \caption{Overview of our proposed model.}
    \label{fig:model_framework}
\end{figure*}

\subsection{Task Definition}
Our goal is to achieve zero-shot or few-shot translation from $T$ low-resource languages $L_1, L_2, ..., L_T$ to the target language $L_y$ with the help of a particular high-resource language $\widehat{L}$. For the high-resource language $\widehat{L}$, there are triples of data $\mathcal{D}_{\widehat{L}}=\{(\mathbf{i}, \mathbf{x}, \mathbf{y})\}$, where $\mathbf{i}$ is the image and $\mathbf{x}$ and $\mathbf{y}$ are the descriptions in $\widehat{L}$ and $L_y$ respectively. For each low-resource language $L_i$, only paired data $\mathcal{D}_{L_i}=\{(\mathbf{i}, \mathbf{x})\}$ are available. Note that different languages never share the same images.


\subsection{Model Framework}
\label{model_framework}
As shown in Figure \ref{fig:model_framework}, our model consists of four sub-modules: \emph{image encoder}, \emph{source encoder}, \emph{target decoder} and \emph{contrastive module}. 

We use Vision Transformer (ViT) \citep{vit} as the \emph{image encoder} to extract visual features. ViT first splits the image into several patches, and then feed the sequence of embed patches with a special \texttt{[class]} token into Transformer \citep{transformer}. Finally, the image is encoded as a sequence of vectors $\mathbf{v}=(v_0, v_1, ..., v_m)$, where $v_0$ is the representation of \texttt{[class]} token which can be regarded as the global representation of the image, and $\mathbf{v}^p=(v_1, ..., v_m)$ are the patch-level representations. In next sections, we use $v_0$ for sentence-level contrastive learning and $\mathbf{v}^p$ for token-level contrastive learning.

The \emph{source encoder} consists of $N$ Transformer encoder layers, which is shared across all languages ($L_{1...T}$ and $\widehat{L}$). For the input sentence $\mathbf{x}=(x_1, ..., x_n)$, the output of \emph{source encoder} is denoted as $\mathbf{w}=(w_1, ..., w_n)$. The \emph{target decoder} consists of $N$ Transformer decoder layers. For the sentence pairs $(\mathbf{x}, \mathbf{y})$, the cross-entropy loss is defined as:
\begin{equation}
    \mathcal{L}_{\rm CE}=-\sum_{i=1}^{|\mathbf{y}|}\log p(y_i^*|\mathbf{y}_{<i}, \mathbf{x}).
\end{equation}

The \emph{contrastive module} aims to align the output of \emph{image encoder} and \emph{source encoder}, which contains both sentence-level and token-level parts. We will introduce them in Section \ref{section:sctr} and \ref{sec:tctr}.




\subsection{Sentence-level Contrastive Learning}
\label{section:sctr}

We start with the sentence-level contrastive learning objective, which aims at learning coarse alignment between image and text.

\paragraph{Contrastive Learning}
The idea of contrastive learning \citep{NIPS2016_6b180037} is to make the representations of corresponding pairs closer and, on the contrary, to make the irrelevant pairs farther.

Given two sets $\mathbf{X}=\{x_i\}_{i=1}^M$ and $\mathbf{Y}=\{y_i\}_{i=1}^M$, for each $x_i$, the positive example is $(x_i, y_i)$ and the remaining $M-1$ irrelevant pairs $(x_i, y_j) (i\neq j)$ are considered as negative examples. The contrastive loss between $\mathbf{X}$ and $\mathbf{Y}$ is defined as:
\begin{equation}
\begin{aligned}
    \mathcal{L}_{\rm ctr}(\mathbf{X},\mathbf{Y})=-\sum_{i=1}^M\log\frac{ \exp(s(x_{i},y_{i})/\tau)}{\sum_{j=1}^M\exp(s(x_{i},y_{j})/\tau)},\label{eq:infonce}
\end{aligned}
\end{equation}
where $s()$ is the cosine similarity function $s(a,b)=a^\top b/\|a\|\|b\|$. $\tau$ is the temperature hyper-parameter to control the strength of penalties on hard negative samples \citep{Wang_2021_CVPR}.

\paragraph{Sentence-level Contrast}
Sentence-level contrastive learning aims to align the sentence-level representations across modalities, which are defined as follows:
\begin{align}
   & w^s = \frac{1}{n}\sum_{i=1}^n w_i, \\
   & v^s = v_0 \label{eq:v_emb}.
\end{align}

We then calculate the contrastive loss within a batch of size $B$, whose textual representations and visual representations are $\mathbf{W}^s=\{w_1^s,...w_B^s\}$ and $\mathbf{V}^s=\{v_1^s,...,v_B^s\}$, respectively. The corresponding pairs of images and captions $(w_i^s, v_i^s)$ are positive examples, and other pairs $(w_i^s, v_j^s) (i\neq j)$ are considered as negative examples. Finally, the loss function of sentence-level contrastive learning is defined as follows:
\begin{equation}
\begin{aligned}
    \mathcal{L}_{\rm s-ctr}(\mathbf{W}^s,\mathbf{V}^s)=\mathcal{L}_{\rm ctr}(\mathbf{W}^s,\mathbf{V}^s) + \mathcal{L}_{\rm ctr}(\mathbf{V}^s,\mathbf{W}^s).\label{s-ctr}
\end{aligned}
\end{equation}

Since we have image-text pairs in different languages within a batch, we first separate the batch into several mini-batches according to the language, and then calculate the contrastive loss for every language respectively. It is worth mentioning that we also calculate contrastive loss for target language $L_y$ with paired data $\{(\mathbf{i}, \mathbf{y})\}$ in $D_{\widehat{L}}$. We will analyze its effect in Section \ref{analysis_con}.



\subsection{Token-level Contrastive Learning}
\label{sec:tctr}
Though sentence-level contrastive learning can learn coarse-grained alignment between modalities, it may ignore some detailed information, which is crucial for predicting translations. To achieve better alignment between modalities, we propose token-level contrastive learning to learn fine-grained correspondences between images and text.


\paragraph{Selective Attention}  
To model the correlations between image patches and words, we use selective attention \citep{li-etal-2022-vision} to learn the patch-level contribution of images. For patch-level visual representations $\mathbf{v}^p=(v_1,...v_m)$ and word-level textual representations $\mathbf{w}=(w_1, ..., w_n)$, the query, key and value of selective attention are $\mathbf{w}, \mathbf{v}^p, \mathbf{v}^p$, respectively:
\begin{equation}
    \mathbf{v}^t=\text{Softmax}\left(\frac{(W_Q\cdot\mathbf{w})(W_K\cdot\mathbf{v}^p)^\top}{\sqrt{d_k}}\right)(W_V\cdot\mathbf{v}^p),
\end{equation}
where $W_Q$, $W_K$ and $W_V$ are learnable matrix parameters. 

\paragraph{Token-level Contrast}
After the selective attention, we obtain two sequences $\mathbf{w}=(w_1,...,w_n)$ and $\mathbf{v}^t=(v_1^t,...,v_n^t)$ with the same length of $n$. We then calculate the token-level contrastive loss within each pair of sequences. Tokens with same index $(w_i, v_i^t)$ are positive examples, and other pairs of tokens $(w_i, v_j^t) (i\neq j)$ are negative examples.
The token-level contrastive loss is as follows:
\begin{equation}
\begin{aligned}
    \mathcal{L}_{\rm t-ctr}(\mathbf{w}, \mathbf{v}^t)=\mathcal{L}_{\rm ctr}(\mathbf{w}, \mathbf{v}^t) + \mathcal{L}_{\rm ctr}(\mathbf{v}^t, \mathbf{w}).
    \label{eq:l_tctr}
\end{aligned}
\end{equation}
The token-level contrastive loss of all image-text pairs will be summed together.

\subsection{Coarse-to-fine Training Strategy}
To combine sentence-level and token-level objectives together, we propose a 2-stage \emph{coarse-to-fine training strategy}, the intuition behind which is to first learn coarse-grained alignment through the sentence-level objective, and then add fine-grained alignment with the token-level objective.

\paragraph{Stage 1}
For the first stage of training, the model is trained with cross-entropy loss of the high-resource language $\widehat{L}$ and sentence-level contrastive loss of all languages (including target language $L_y$):
\begin{equation}
\begin{aligned}
    \mathcal{L}_{\rm coarse}&=\mathbb{E}_{(\mathbf{x}, \mathbf{y})\in\mathcal{D}_{\widehat{L}}}\mathcal{L}_{\rm CE}(\mathbf{x}, \mathbf{y}) \\
    &+ \lambda_s\mathbb{E}_{(\mathbf{i}, \mathbf{y})\in\mathcal{D}_{\widehat{L}}}\mathcal{L}_{\rm s-ctr}(\mathbf{i}, \mathbf{y}) \\
    &+ \lambda_s\mathbb{E}_{(\mathbf{i}, \mathbf{x})\in\mathcal{D}_{\widehat{L}}}\mathcal{L}_{\rm s-ctr}(\mathbf{i}, \mathbf{x}) \\
    &+ \lambda_s\sum_{i=1}^{T}\mathbb{E}_{(\mathbf{i}, \mathbf{x})\in\mathcal{D}_{L_i}}\mathcal{L}_{\rm s-ctr}(\mathbf{i}, \mathbf{x}),
    \label{sctr}
\end{aligned}
\end{equation}
where $\lambda_s$ is the weight hyper-parameter of sentence-level contrastive loss.

\paragraph{Stage 2}
For the second stage of training, we add the token-level contrastive loss to Eq. \ref{sctr}, which can be formulated as follows:
\begin{equation}
\begin{aligned}
\mathcal{L}_{\rm fine}&=\mathcal{L}_{\rm coarse} \\
&+ \lambda_t\mathbb{E}_{(\mathbf{i}, \mathbf{y})\in\mathcal{D}_{\widehat{L}}} \mathcal{L}_{\rm t-ctr}(\mathbf{i}, \mathbf{y}) \\
&+ \lambda_t\mathbb{E}_{(\mathbf{i}, \mathbf{x})\in\mathcal{D}_{\widehat{L}}} \mathcal{L}_{\rm t-ctr}(\mathbf{i},\mathbf{x}) \\
    &+ \lambda_t\sum_{i=1}^{T}\mathbb{E}_{(\mathbf{i}, \mathbf{x})\in\mathcal{D}_{L_i}}\mathcal{L}_{\rm t-ctr}(\mathbf{i},\mathbf{x}),
\end{aligned}
\label{tctr}
\end{equation}
where $\lambda_t$ is the weight hyper-parameter of token-level contrastive loss.

\paragraph{Zero-shot and Few-shot Translation}
After 2-stage training with contrastive loss, we can directly evaluate the performance of the trained model on zero-shot translation. Furthermore, we can use small amount of additional parallel data of low-resource languages $\mathcal{D}_L=\{(\mathbf{x}, \mathbf{y})\}$ to finetune the model, and then evaluate the performance on few-shot translation. During finetuning, only cross-entropy loss is used:
\begin{equation}
\mathcal{L}_{\rm finetune}=\mathbb{E}_{(\mathbf{x}, \mathbf{y})\in\mathcal{D}_{L}}\mathcal{L}_{\rm CE}(\mathbf{x}, \mathbf{y}).
\end{equation}

\begin{table}[!t]
\centering
\small
\begin{tabular}{lcccc}
\toprule
Directions & Multi30K & MsCOCO & VizWiz & Total  \\ \midrule
DE$\rightarrow$EN   & 10,000   & 40,000 & 10,136 & 60,136 \\
FR$\rightarrow$EN   & 10,000   & 40,000 & 10,136 & 60,136 \\
CS$\rightarrow$EN   & 9,000    & 41,000 & 10,136 & 60,136 \\ \bottomrule
\end{tabular}
\caption{Detailed dataset statistics.}
\label{table:dataset}
\end{table}

\section{Experiments}
\begin{table*}[tb]
\centering
\begin{tabular}{l|ccc|c|c}
\toprule
\multirow{2}{*}{\textbf{Models}} & \multicolumn{3}{c|}{\textbf{FR$\rightarrow$EN}}              & \textbf{CS$\rightarrow$EN} & \multirow{2}{*}{\textbf{Average}} \\
                                 & Test2016       & Test2017       & MsCOCO         & Test2016       &                                   \\\midrule
Baseline                          & 0.30           & 0.14           & 0.29           & 0.09           & 0.21                              \\
S-CTR                            & 8.95           & 7.88           & 9.32           & 7.23           & 8.35                              \\
S+T-CTR                          & \textbf{17.76} & \textbf{14.74} & \textbf{16.97} & \textbf{13.58} & \textbf{15.76}                    \\ \bottomrule
\end{tabular}
\caption{BLEU scores of FR$\rightarrow$EN and CS$\rightarrow$EN on zero-shot translation. }
\label{table:zero-shot}
\end{table*}
\begin{figure*}[tb]
    \centering
    \small
    \subfigure[FR$\rightarrow$EN Test2016]{
        \begin{minipage}{0.48\linewidth}
            \centering
            \includegraphics[scale=0.25]{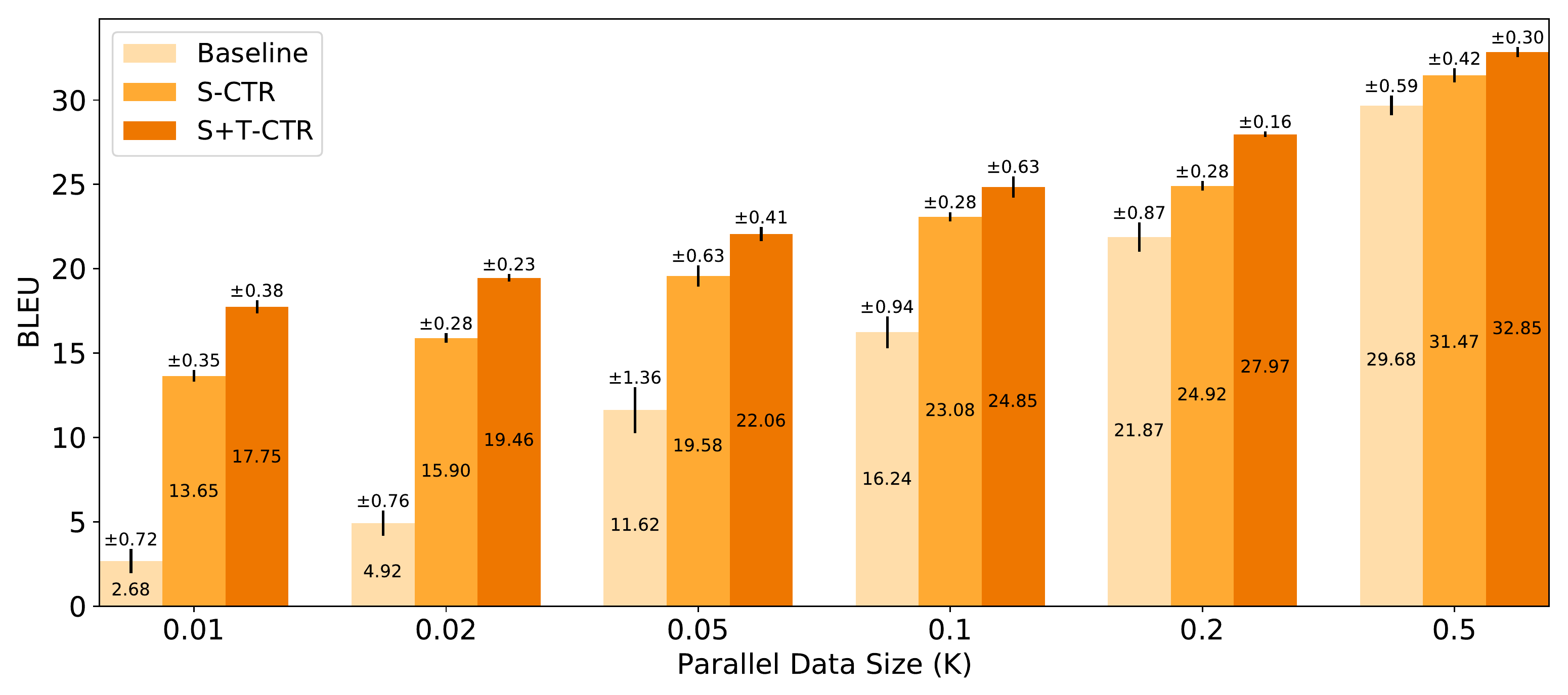}
        \end{minipage}
    }
    \subfigure[FR$\rightarrow$EN Test2017]{
        \begin{minipage}{0.48\linewidth}
            \centering
            \includegraphics[scale=0.25]{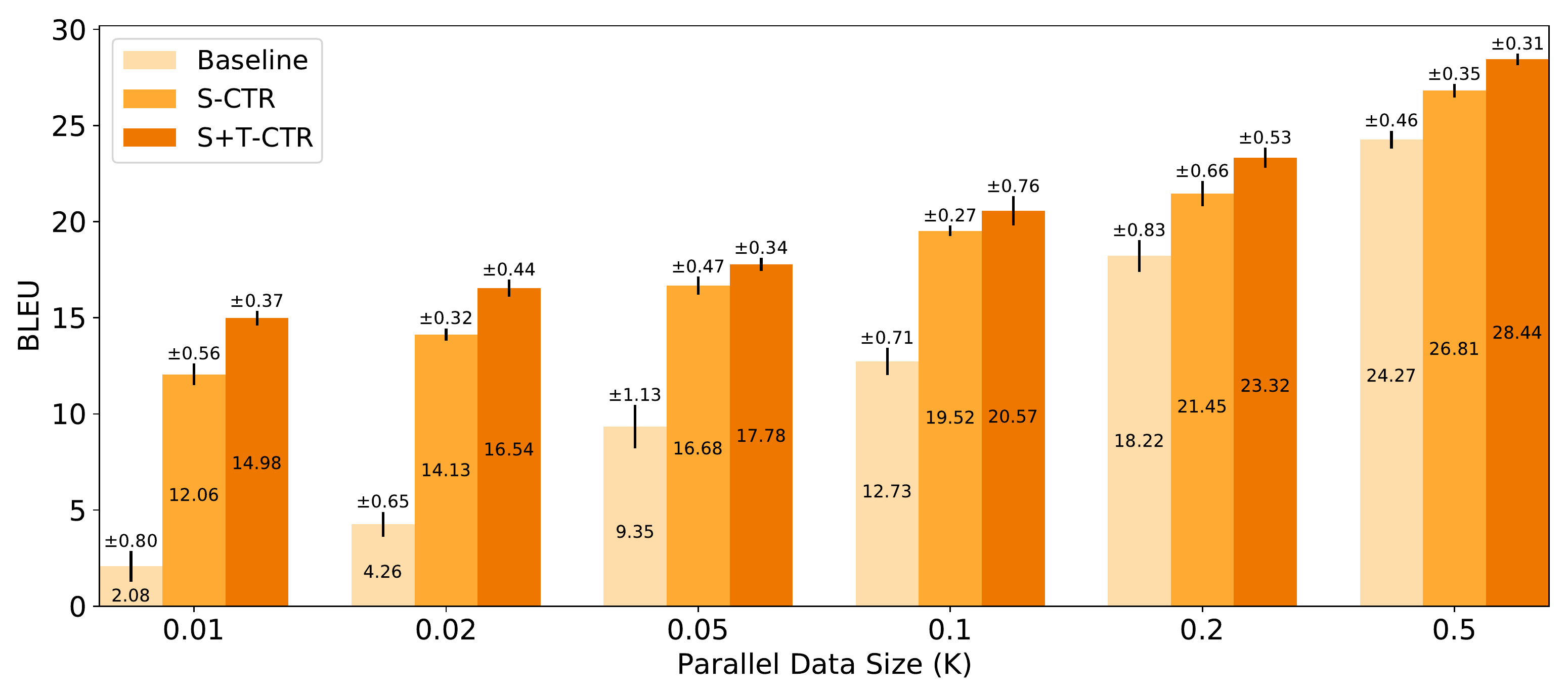}
        \end{minipage}
    }
    \subfigure[FR$\rightarrow$EN MsCOCO]{
        \begin{minipage}{0.48\linewidth}
            \centering
            \includegraphics[scale=0.25]{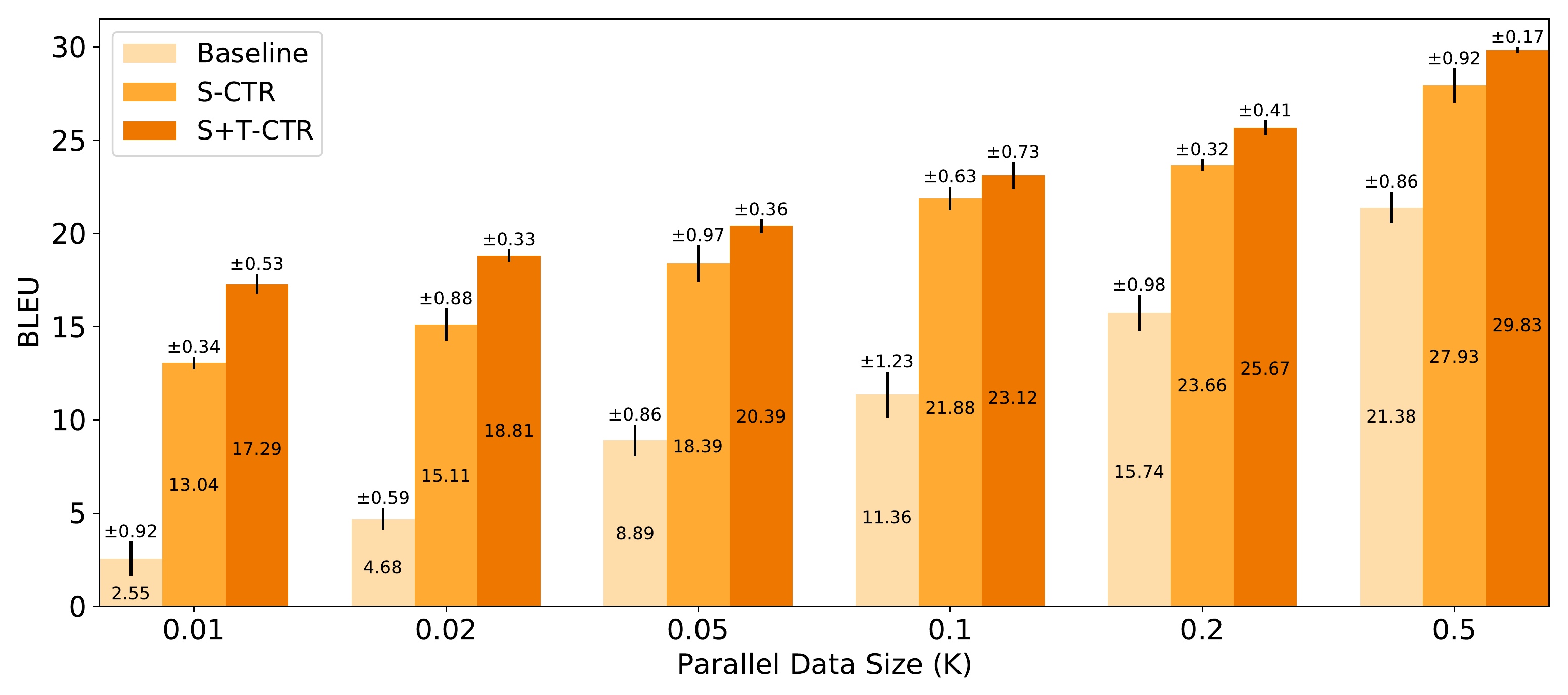}
        \end{minipage}
    }
    \subfigure[CS$\rightarrow$EN Test2016]{
        \begin{minipage}{0.48\linewidth}
            \centering
            \includegraphics[scale=0.25]{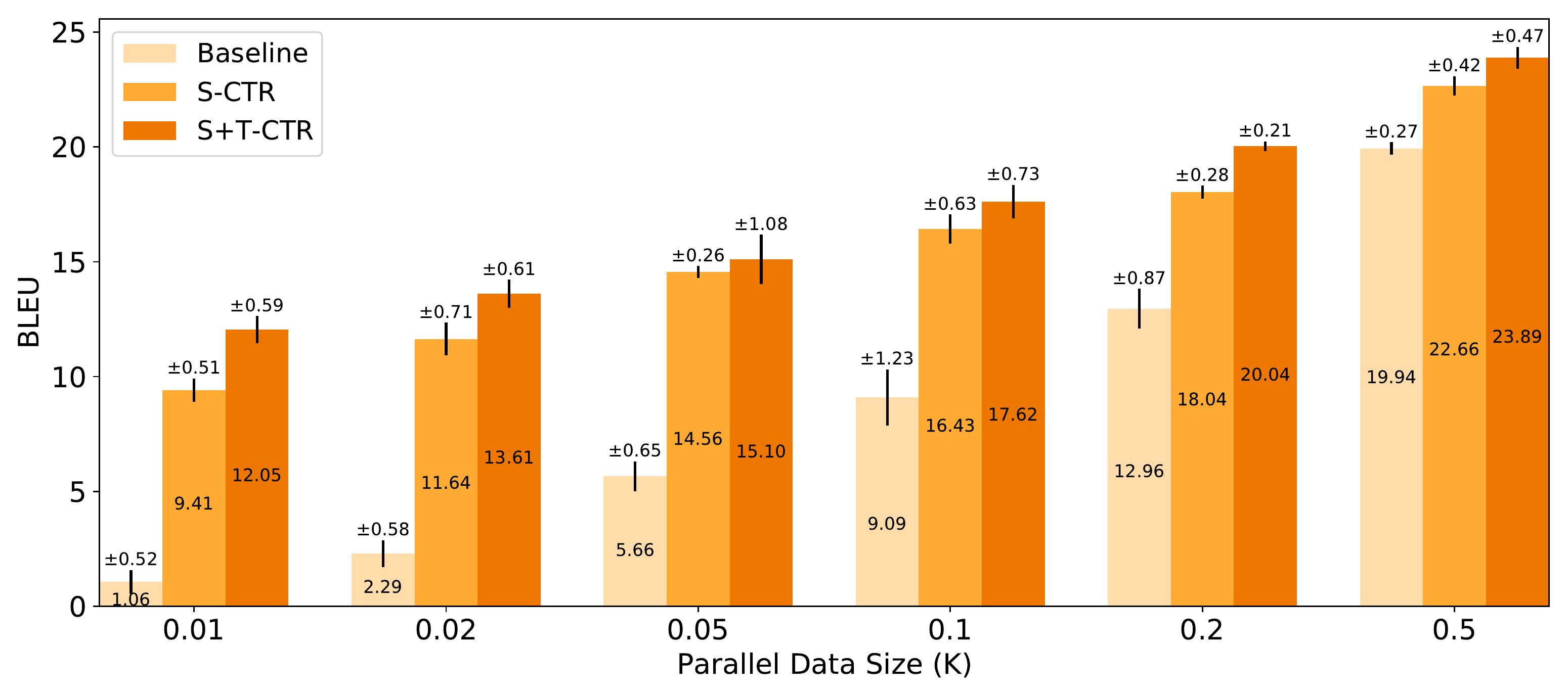}
        \end{minipage}
    }
    \caption{BLEU scores of FR$\rightarrow$EN and CS$\rightarrow$EN on few-shot translation.}
    \label{fig:FRTest2016}
\end{figure*}

\begin{figure*}[tb]
    \centering
    \includegraphics[width=\linewidth]{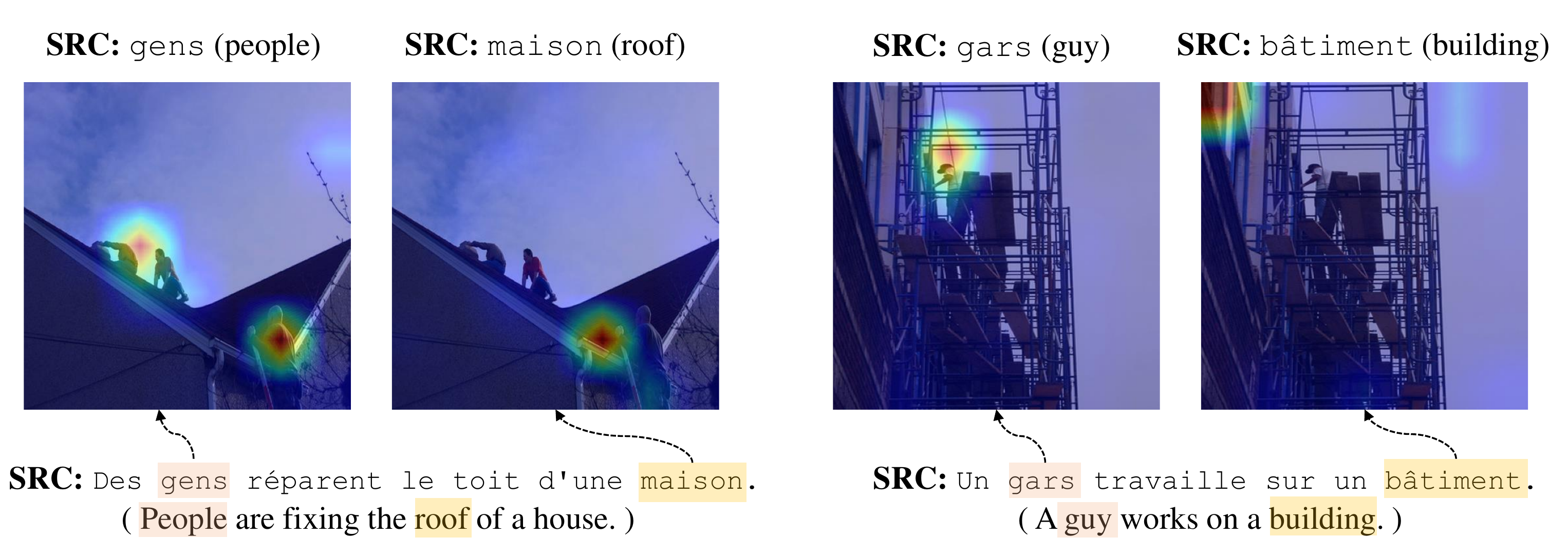}
    \caption{Attention maps of the selective attention module of two cases.}
    \label{fig:attn_map}
\end{figure*}

\subsection{Datasets}
In our experiments, we select German (DE) as the high-resource language and English (EN) as the target language. We choose French (FR) and Czech (CS) as two low-resource languages and test the performance of FR$\rightarrow$EN and CS$\rightarrow$EN on zero-shot and few-shot translation. Due to the scarcity of image-text pairs in German, French, and Czech, we create pseudo data with machine translation models from two image captioning datasets in English.
\paragraph{Multi30K}
Multi30K \citep{DBLP:conf/acl/ElliottFSS16} dataset contains images with annotations in four languages: English, German, French, and Czech. The training and validation sets consist of 29,000 and 1,014 instances, respectively. We evaluate our model on Test2016, Test2017, and MsCOCO test sets, which contain 1,000, 1,000, and 456\footnote{5 sentences are removed because they appear in the MsCOCO dataset, which is part of our training set.} instances. For Czech$\rightarrow$English task, only Test2016 is available.
\paragraph{MsCOCO}
MsCOCO \citep{DBLP:conf/eccv/LinMBHPRDZ14} dataset contains images with English captions. We use the Captioning 2015 set for our experiments. After filtering out the unannotated images, there are 121,000 image-text pairs in total.
\paragraph{VizWiz}
VizWiz \citep{DBLP:conf/eccv/GurariZZB20} dataset also contains images with English captions. There are 30,408 image-text pairs in total. 
\paragraph{Pseudo Data}
Since the MsCOCO and VizWiz datasets only have English captions of images. We use pretrained machine translation models to translate English captions into German, French and Czech. The detailed information of the machine translation models can be seen in Appendix \ref{qpt}. 

\paragraph{Dataset Composition}
After creating the pseudo data, we divide the above three datasets into three equal parts for DE$\rightarrow$EN, FR$\rightarrow$EN, and CS$\rightarrow$EN, respectively. As shown in Table \ref{table:dataset}, each source language has 60,136 image-text pairs with annotations in its own language, which are used for cross-modal contrastive learning. At the same time, the 60,136 German$\rightarrow$English sentence pairs are used for training of translation task. All sentences are segmented into subword units using byte-pair encoding (BPE) \citep{sennrich-etal-2016-neural}. The vocabulary is shared for all source languages and the target language, with a size of 18K.




\subsection{System Settings} 
We use vision transformer in pre-trained CLIP \citep{clip} model as the \emph{image encoder}. The patch size is 16$\times$16, and the resolution size is 224. The sequence length is 50, which contains a special \texttt{[class]} token and 49 feature tokens. The \emph{source encoder} and \emph{target decoder} are based on Transformer \citep{transformer} architecture. Both the encoder and decoder have $N=6$ layers. The number of attention heads is set to 4. The dropout is set to 0.3, and the value of label smoothing is 0.1. For training, we use Adam optimizer \citep{adam} and 2000 warm-up updates. The learning rate is 5e-4. Each batch contains up to 16K tokens. We train the model for up to 70 epochs. For our 2-stage training strategy, the first half of training is Stage 1, and the rest is Stage 2.

For sentence-level contrastive learning, the temperature hyper-parameter $\tau_{s}$ is set to 0.007 and the weight hyper-parameter $\lambda_{s}$ is set to 5. For token-level contrastive learning, $\tau_{t}$ is 0.1 and $\lambda_{t}$ is 1.

For evaluation, we average the last 5 checkpoints and use beam search with a beam size of 5. We use sacreBLEU\footnote{\url{https://github.com/mjpost/sacrebleu}} \citep{post} to compute the BLEU \citep{Bleu} scores on detokenized instances\footnote{sacreBLEU signature: nrefs:1 | bs:1000 | seed:12345 | case:lc | eff:no | tok:13a | smooth:exp | version:2.0.0}.
For few-shot translation, we randomly sample 5 groups of parallel data from the training set of Multi30K and report the means and standard deviations. All experiments are done on 4 TITAN Xp GPUs. We implement our system based on \emph{fairseq}\footnote{\url{https://github.com/pytorch/fairseq}} \citep{fairseq}.

\subsection{Baseline Systems}
Our baseline is text-only Transformer trained with DE$\rightarrow$EN sentence pairs. For zero-shot translation, we directly evaluate the baseline model. For few-shot translation, we finetune the baseline model with the same parallel corpus in low-resource languages as our model. All the configurations of the baseline are the same as our model. 

\subsection{Results}
We evaluate the baseline, our model with only sentence-level contrastive loss (\textbf{S-CTR}), and our model with both sentence-level and token-level contrastive loss (\textbf{S+T-CTR}) under zero-shot and few-shot scenarios. 
\paragraph{Zero-shot Translation}
Table \ref{table:zero-shot} shows the results on zero-shot translation. The baseline without contrastive learning does not have the capability of zero-shot translation. 
On the contrary, S-CTR and S+T-CTR gain significantly improvements over the baseline. Compared with S-CTR, the S+T-CTR model has a further improvement of 7.41 BLEU score on average, which proves that more fine-grained alignment can significantly improve the performance on zero-shot machine translation.

\paragraph{Few-shot Translation}
Figure \ref{fig:FRTest2016} shows the results on few-shot translation on four test sets. The S+T-CTR model consistently outperforms the baseline and the S-CTR model under different amounts of parallel data, demonstrating the effectiveness of our method in few-shot scenarios.

\section{Analysis}
\begin{figure}[t]
    \centering
    \subfigure[Baseline]{
        \begin{minipage}{0.46\linewidth}
            \centering
            \includegraphics[scale=0.25]{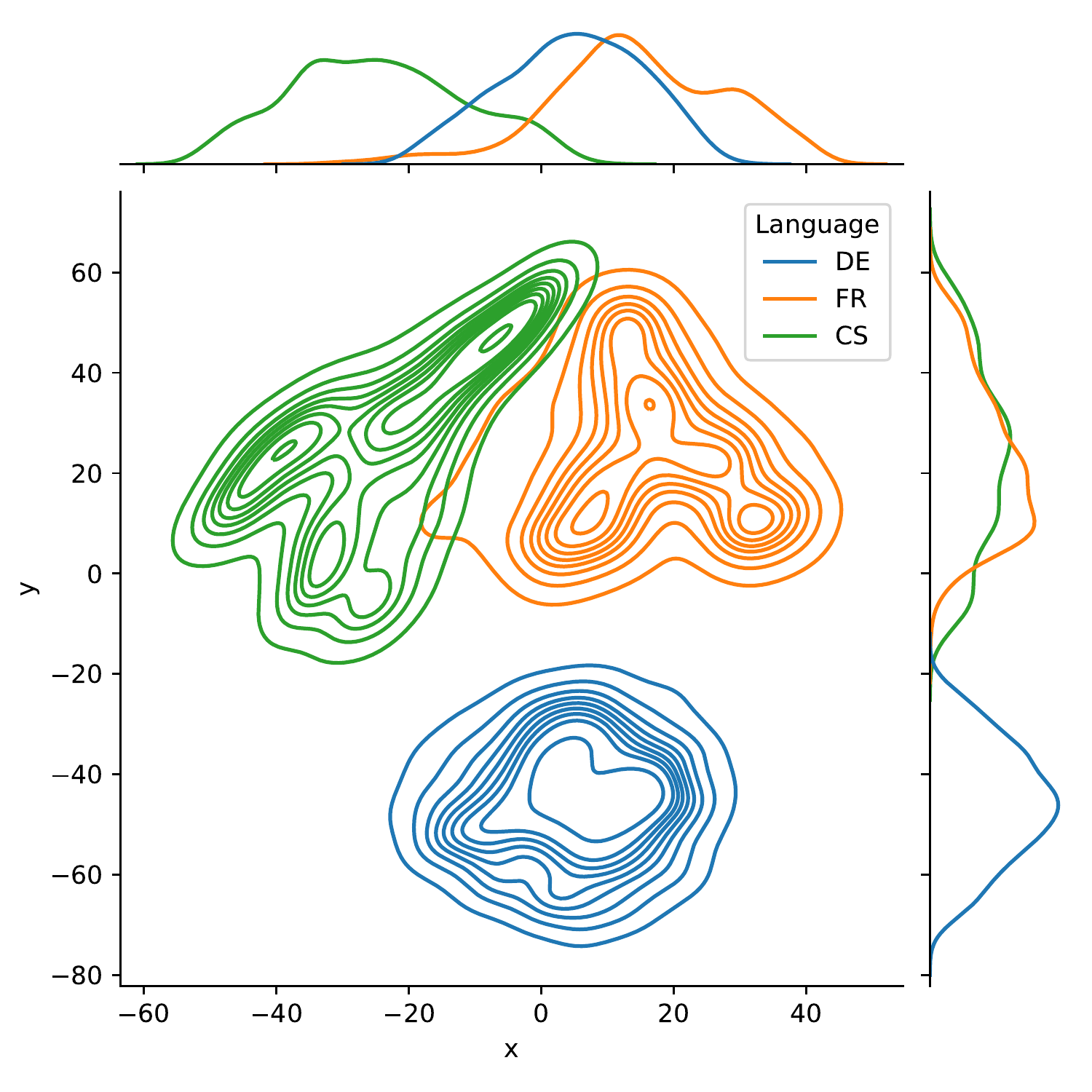}
        \end{minipage}
    }
    \subfigure[S+T-CTR model]{
        \begin{minipage}{0.46\linewidth}
            \centering
            \includegraphics[scale=0.25]{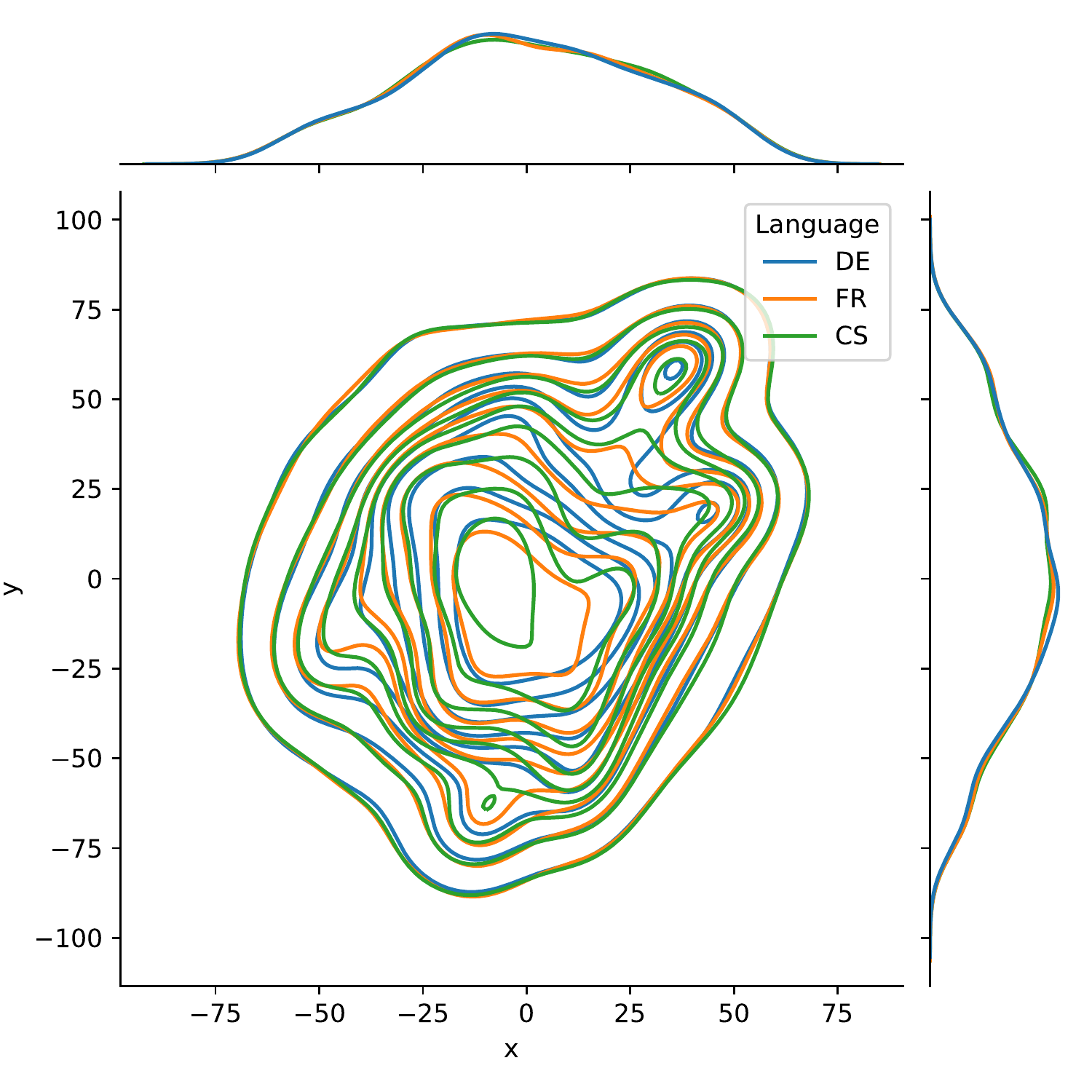}
        \end{minipage}
    }
    \caption{Visualization of source representations for DE, FR, and CS under the zero-shot scenario. (a) baseline. (b) S+T-CTR model. Sentences are from Multi30K Test2016 sets of DE$\rightarrow$EN, FR$\rightarrow$EN, and CS$\rightarrow$EN.}
    \label{fig:cross_lan}
\end{figure}

\begin{table}[!t]
	\centering
	\begin{tabular}{lccc}
		\toprule  
		\textbf{Models} & \textbf{R@1$\uparrow$} & \textbf{R@5$\uparrow$} &  \textbf{R@10$\uparrow$}\\
		\midrule
		Baseline & 0.2 & 0.8 & 1.3 \\
        S-CTR & 34.4 & 65.0 & 75.4 \\
        S+T-CTR & \textbf{36.3} & \textbf{66.5} & \textbf{76.1} \\
		\bottomrule  
	\end{tabular}
	\caption{Text-to-image retrieval on FR$\rightarrow$EN Test2016.}
	\label{image acc}
\end{table}

\subsection{Cross-modal Alignment}
\label{cross-modal}
The main idea of our method is to align multilingual text and images in their representation space. To verify this alignment, we conduct the text-to-image retrieval experiment and visualize the attention map of the selective attention module. 

\paragraph{Text-to-image Retrieval}
Text-to-image retrieval means finding the top-$K$ nearest images to the text. We compute the Recall@$K$ score for $K=1,5,10$. As shown in Table \ref{image acc}, S-CTR gains a substantial 34.2/64.2/74.1\% increase in R@1/5/10 over the baseline, which proves the effectiveness of contrastive learning for cross-modal alignment. In addition, S+T-CTR gains an extra 1.9/1.5/0.7\% increase in R@1/5/10, proving that the fine-grained learning objective enables better alignment.

\paragraph{Attention Maps}
To further verify the effect of token-level contrastive learning for cross-modal alignment, we extract attention maps of the selective attention module. Figure \ref{fig:attn_map} demonstrates that the selective attention module successfully notices the semantically related areas. For example, the French word "gens" (means "people") corresponds to the three people and the word "maison" (means "roof") corresponds to the roof area.


\subsection{Cross-lingual Alignment}
Section \ref{cross-modal} analyses the effectiveness of contrastive learning on cross-modal alignment. However, our ultimate goal is to achieve cross-lingual alignment through cross-modal alignment, which means to learn a shared space for all languages.



To analyze, we compare the baseline and S+T-CTR model under the zero-shot scenario, which means no FR$\rightarrow$EN or CS$\rightarrow$EN parallel data is available. We average the output of the \emph{source encoder} and use T-SNE \cite{2008Visualizing} to reduce the dimension into two for visualization. As shown in Figure \ref{fig:cross_lan}, without contrastive learning, there is a clear distinction between different source languages. On the contrary, with contrastive learning, the representations of three languages have obviously overlapped, which proves that our method learned good cross-lingual alignment.

\subsection{Ablation Studies}
\paragraph{Target Language Contrast}
\label{analysis_con}
The target language is generally isolated from the source language in standard machine translation. However, we found that adding the target language into contrastive learning is effective.
As shown in Table \ref{table:target_ablation}, models without contrastive learning of the target language have a significant drop in BLEU score under both zero-shot and few-shot situations. We conclude that contrastive learning of the target language can help establish connections between source and target languages, which will be beneficial for translation.

\begin{table}[!t]
\small
\centering
\resizebox{\linewidth}{!}{
\begin{tabular}{c|c|cc|cc}
\toprule
\multirow{2}{*}{\textbf{Models}}  & \multirow{2}{*}{\textbf{Target}} & \multicolumn{2}{c|}{\textbf{FR$\rightarrow$EN}} & \multicolumn{2}{c}{\textbf{CS$\rightarrow$EN}} \\
                         &                         & ZS           & FS100       & ZS        & FS100         \\ \midrule
Baseline                 & -                       & 0.24         & 13.44       & 0.09      & 9.10           \\ \midrule
\multirow{2}{*}{S-CTR}   & $\times$                        & 7.81         & 12.55           & 6.93      & 8.67             \\
                         &   $\checkmark$                      & 8.71         & 21.49       & 7.23      & 16.43         \\ \midrule
\multirow{2}{*}{S+T-CTR} &  $\times$                       & 14.47        & 13.16           & 10.97     & 8.24    \\
                         & $\checkmark$                        & \textbf{16.49}        & \textbf{22.85}       & \textbf{13.58}     & \textbf{17.62}         \\ \bottomrule
\end{tabular}}
\caption{Ablation study on contrastive learning of the target language. ZS means zero-shot translation, FS100 means few-shot translation with 100 parallel sentences.}
\label{table:target_ablation}
\end{table}

\paragraph{Contrastive Loss vs. L2 Loss}
Contrastive loss is not the only way to draw the distance between modalities. We try to replace the contrastive loss with L2 loss:
\begin{equation}
\mathcal{L}_{L2}=\sum_{i=1}^M\|x_i-y_i\|^2.
\end{equation}
As shown in Table \ref{table:l2loss}, the contrastive loss performs better than the L2 loss. We believe it is because the contrastive loss can not only bring the corresponding pairs closer but also push the irrelevant pairs farther with negative examples.

\begin{table}[!t]
\centering
\small
\resizebox{\linewidth}{!}{
\begin{tabular}{c|c|cc|cc}
\toprule  
    \multirow{2}{*}{\textbf{Models}}    & \multirow{2}{*}{\textbf{Loss}} & \multicolumn{2}{c|}{\textbf{FR$\rightarrow$EN}}      & \multicolumn{2}{c}{\textbf{CS$\rightarrow$EN}}       \\ 

                           &                       & ZS             & FS100          & ZS             & FS100          \\ 
\midrule
Baseline                   & -                     & 0.24           & 13.44          & 0.09           & 9.10            \\ \midrule
\multirow{2}{*}{S-level}   & L2                    & 8.45              & 19.60              & 6.83             & 15.05              \\
                           & CTR                   & 8.71           & 21.49          & 7.23           & 16.43          \\ \midrule
\multirow{2}{*}{S+T-level} & L2                    & 6.02              & 15.61              &  5.94              & 12.76     \\
                           & CTR                   & \textbf{16.49} & \textbf{22.85} & \textbf{13.58} & \textbf{17.62} \\ 
\bottomrule
\end{tabular}}
\caption{BLEU scores of models with L2 loss and contrastive loss. ZS means zero-shot translation, FS100 means few-shot translation with 100 parallel sentences.}
\label{table:l2loss}
\end{table}
\begin{figure}[!t]
\centering
    \subfigure[Sentence-level]{
        \begin{minipage}{0.45\linewidth}
            \small
            \centering
            \includegraphics[scale=0.25]{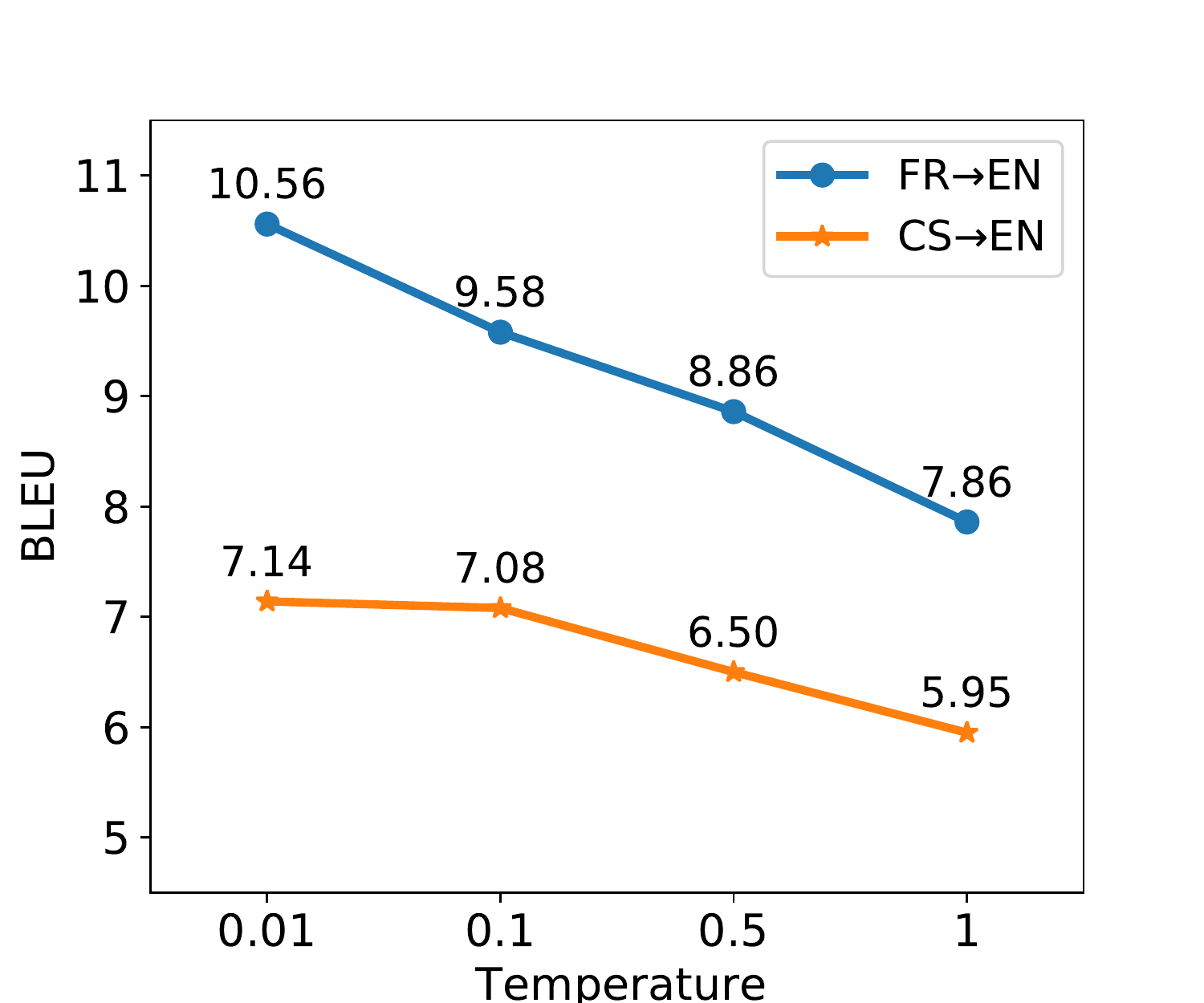}
        \end{minipage}
    }
    \subfigure[Token-level]{
        \begin{minipage}{0.45\linewidth}
            \small
            \centering
            \includegraphics[scale=0.25]{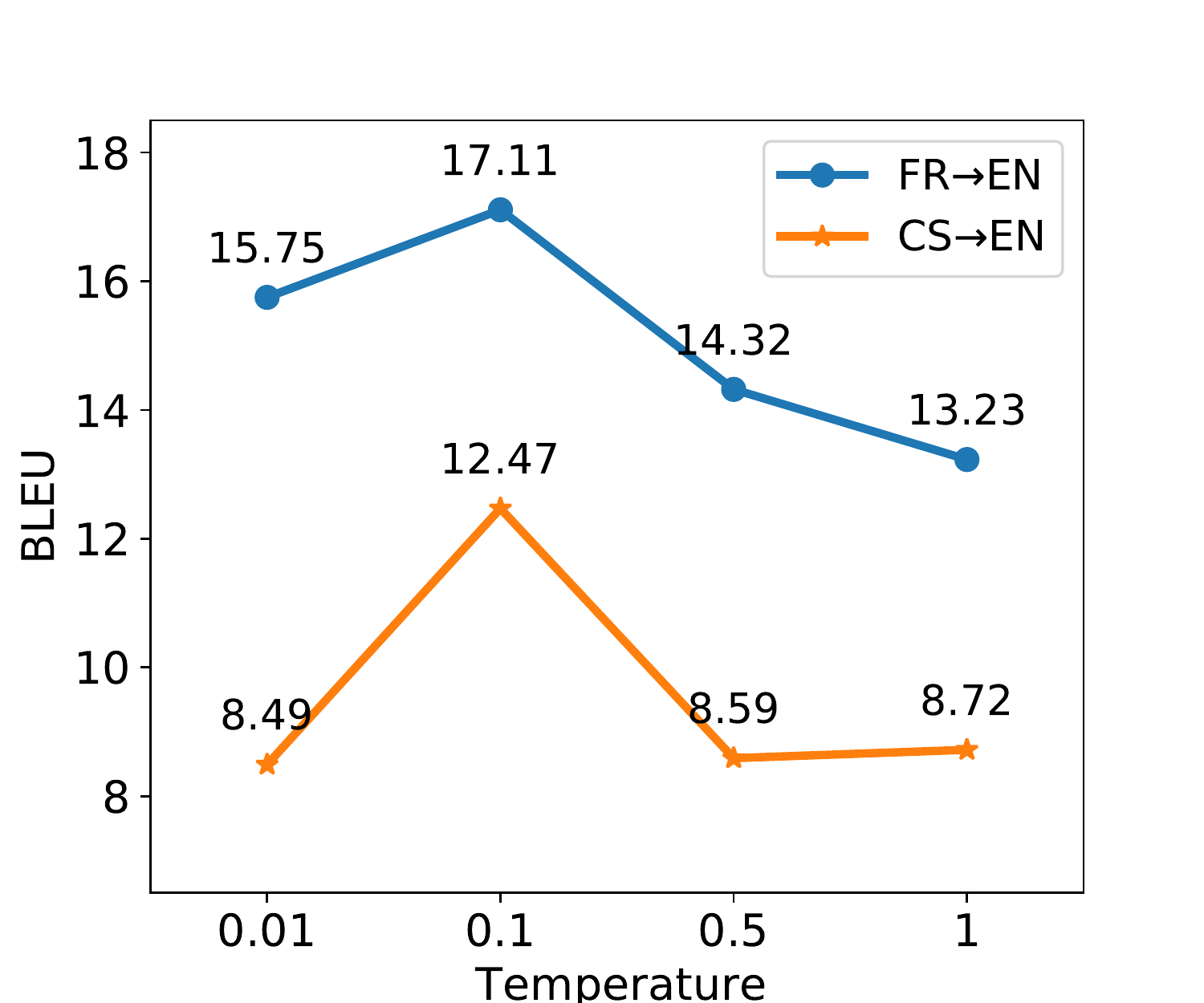}
        \end{minipage}
    }
    \caption{BLEU scores on Multi30K validation set against different temperatures for sentence-level and token-level contrastive learning.}
    \label{fig:tem}
\end{figure}
\subsection{Temperature Hyper-parameter}
The temperature $\tau$ is an important hyper-parameter in contrastive learning. A lower temperature can help the model distinguish positive example from negative ones. Here we choose 0.01, 0.1, 0.5 and 1 for experiments. Figure \ref{fig:tem} shows the BLEU scores against different temperatures on validation set.

For sentence-level contrastive learning, we observe that lower temperatures obtain better results. We try to choose the temperature as low as possible. However, a temperature lower than 0.007 may lead to gradient explosion. So we finally select $\tau_s=0.007$. For token-level contrastive learning, we found $\tau_t=0.1$ achieved best results on validation set. We think it is because that tokens in a sentence should not be excessively distinguished.

\begin{table*}[!t]
\small
\centering
\begin{tabular}{ccl}
\toprule
\multicolumn{2}{l|}{\textbf{Models}}                                             &                                                                       \\ \midrule
\multicolumn{3}{c}{\textbf{Case 1 FR$\rightarrow$EN}}                                                                                                    \\ \midrule
\multicolumn{1}{c|}{}                       & \multicolumn{1}{c|}{SRC} & Des enfants sont dehors , jouant dans la terre à côté de deux arbres. \\
\multicolumn{1}{c|}{\multirow{-2}{*}{Ref.}} & \multicolumn{1}{c|}{TGT} & Some children are outside playing in the dirt where two trees are.    \\ \midrule
\multicolumn{2}{l|}{S-CTR (ZS)}                                          &  The young children are {\color[HTML]{FE0000}{playing a game of dirt}}. \textit{(two trees)} \\
\multicolumn{2}{l|}{S+T-CTR (ZS)}                                        & The children are outside {\color[HTML]{FE0000}{playing a game of dirt}} next to {\color[HTML]{32CB00}{two trees}}.    \\
\multicolumn{2}{l|}{S+T-CTR (FS100)}                                     & The children are outside {\color[HTML]{32CB00}{playing  a game in the dirt}} near {\color[HTML]{32CB00}{two trees}}.   \\ \midrule
\multicolumn{3}{c}{\textbf{Case 2 CS$\rightarrow$EN}}                                                                                            \\ \midrule
\multicolumn{1}{c|}{}                       & \multicolumn{1}{c|}{SRC} & Muž ve žluté košili a muž v tmavém modrém tričku si povídají.         \\
\multicolumn{1}{c|}{\multirow{-2}{*}{Ref.}} & \multicolumn{1}{c|}{TGT} & A man in a yellow shirt and a man in a dark blue shirt talking.       \\ \midrule
\multicolumn{2}{l|}{S CTR (ZS)}                                          & Man in yellow shirt is {\color[HTML]{FE0000}{crying}}. \textit{(a man in a dark blue shirt)}           \\
\multicolumn{2}{l|}{S+T-CTR (ZS)}                                        & Man in yellow shirt \textit{(and)} a {\color[HTML]{32CB00}{man in a blue shirt}} is {\color[HTML]{FE0000}{smiling}}.  \\
\multicolumn{2}{l|}{S+T-CTR (FS100)}                                     & A man in a yellow shirt and a man {\color[HTML]{32CB00}{in a blue shirt}} is {\color[HTML]{32CB00}{talking}}.         \\ \bottomrule
\end{tabular}
\caption{Qualitative examples from Multi30K Test2016 set. The {\color[HTML]{FE0000}{red text}} indicates the grammar or vocabulary error, \textit{(words in brackets)} indicate the missing words, and the {\color[HTML]{32CB00}{green text}} indicates the correct translations.}
\label{table:case_study}
\end{table*}
\subsection{Case study}
In this section, we make a qualitative analysis with several examples. Table \ref{table:case_study} shows the references and translation results of different models. First, we compare S-CTR and S+T-CTR under the zero-shot scenario. In Case 1, "two trees" have not been translated by the S-CTR model, while the S+T-CTR model translates it correctly. A similar issue occurs in Case 2 (missing "a man in a dark blue shirt"). Both cases suggest that \textbf{fine-grained token-level alignment could avoid missing translation}.

However, both S-CTR and S+T-CTR may have grammar problems under the zero-shot scenario, which can be solved by finetuning with a few parallel data. In Case 1, the phrase "playing a game of dirt" is obviously illogical, while the additional 100 parallel data corrects the preposition "of" to "in", which is more grammatical. This phenomenon shows that \textbf{it is difficult to learn grammar knowledge with contrastive learning, but only a few parallel data can compensate for this}.


\section{Related Work}

\paragraph{Multimodal Machine Translation}
Multimodal Machine Translation aims to introduce visual modality to enhance NMT. Early methods \citep{caglayan2016multimodal, huang-etal-2016-attention, calixto-etal-2016-dcu, delbrouck-dupont-2017-empirical, caglayan-etal-2017-lium, calixto-liu-2017-incorporating,  DBLP:journals/corr/DelbrouckD17, calixto-etal-2017-doubly, libovicky-helcl-2017-attention, caglayan-etal-2018-lium, zhou-etal-2018-visual, helcl-etal-2018-cuni} are mainly based on RNN architecture with attention. Recent methods \citep{ive-etal-2019-distilling, yao-wan-2020-multimodal, yin-etal-2020-novel, liu2021gumbel, 10.1145/3394171.3413715, caglayan-etal-2021-cross, Zhang2020Neural, fang-and-feng-2022-PLUVR, li-etal-2022-vision} based on Transformer further improve the performance. However, recent studies \citep{caglayan-etal-2019-probing, wu-etal-2021-good} found that visual information is often ignored when parallel corpus is sufficient. Therefore, in this paper, we turn to investigate the contribution of visual modality when the parallel corpus is not sufficient.


\paragraph{Zero-shot and Few-shot MT}
Since NMT strongly relies on large scale of parallel data, researchers begin to focus on situations with limited parallel data. Previous methods like unsupervised machine translation \citep{DBLP:conf/emnlp/LampleOCDR18,DBLP:conf/iclr/LampleCDR18,DBLP:conf/iclr/LampleCRDJ18,DBLP:conf/aaai/RenZ00M19,sennrich-zhang-2019-revisiting,ruiter-etal-2019-self} achieve this with abundant monolingual data. Multilingual machine translation \citep{DBLP:conf/naacl/AharoniJF19, DBLP:journals/tacl/LiuGGLEGLZ20, DBLP:conf/emnlp/LinPWQFZL20, pan-etal-2021-contrastive} achieve this with parallel corpus of many other directions. Another line of research is to achieve zero-shot or few-shot translation with the help of visual modality \citep{DBLP:journals/mt/NakayamaN17, DBLP:conf/coling/LiPVK20}, but they failed to achieve satisfactory performance with extremely limited data. We extend this research line and achieve better performance with less data.

\paragraph{Cross-modal Contrastive Learning}
Contrastive learning has lead to a great success in multimodal tasks like cross-lingual transfer \citep{DBLP:conf/naacl/HuangPHNMH21}, video-text understanding \citep{DBLP:conf/emnlp/XuG0OAMZF21}, and so on. One of the most representative methods is CLIP \citep{clip}, which learns good alignment between images and text with contrastive learning. Recent work also shows the power of cross-modal contrastive learning in speech translation \citep{DBLP:journals/corr/abs-2205-02444}. Inspired by these efforts, we propose a cross-modal contrastive learning method to achieve zero-shot and few-shot translation.

\section{Conclusion}
In this paper, we propose a cross-modal contrastive learning method including sentence-level and token-level objectives, which realizes zero-shot and few-shot translation. Experimental results show that our method gains significant improvements over baseline under both scenarios. Further analysis demonstrate that our method learns good cross-modal and cross-lingual alignment. In the future, we will explore how our method enables cross-lingual transfer on more tasks.

\section*{Limitations}

One limitation of our work is the pseudo data we used. Limited by the fact that most of existing image captioning datasets are annotated in English, we have to use additional translation models to generate pseudo captions in German, French and Czech. The lack of real data may impact the performance of our method.


\section*{Acknowledgements}
 The authors would like to thank all the anonymous reviewers for their insightful and valuable comments. 
\bibliography{anthology,custom}
\bibliographystyle{acl_natbib}
\clearpage

\appendix
\section{Translation Models for Pseudo Data}
\label{qpt}
In this section, we introduce the detailed information of translation models we use to construct the pseudo data.
For EN$\rightarrow$DE and EN$\rightarrow$FR directions, we use the pretrained model from \citet{DBLP:conf/wmt/OttEGA18}\footnote{\url{https://github.com/facebookresearch/fairseq/tree/main/examples/translation}}, which consist of 6 encoder and decoder layers. The number of attention heads is set to 16. The dropout is set to 0.3 for EN-DE and 0.1 for EN-FR. The label smoothing is set to 0.1. 

For EN$\rightarrow$CS, we train a Transformer-base model on the WMT2015 EN$\rightarrow$CS training set, which contains about 15M parallel data. The model contains 6 encoder and decoder layers. The number of attention heads is set to 8. The dropout and the label smoothing is set to 0.1. 

We evaluate the EN$\rightarrow$DE and EN$\rightarrow$FR models on the WMT test set \texttt{newstest2014}, and evaluate the EN$\rightarrow$CS model on \texttt{newstest2015}. As shown in Table \ref{table:translation}, the performance of our models is reliable.

\begin{table}[h]
\begin{tabular}{llc}
\toprule
Languge                & Model                    & BLEU \\ \midrule
\multirow{2}{*}{EN$\rightarrow$DE} & \citet{transformer}                &   28.4  \\
                       & \multicolumn{1}{l}{Ours \citep{DBLP:conf/wmt/OttEGA18}} & 29.3  \\ \midrule
\multirow{2}{*}{EN$\rightarrow$FR} & \citet{transformer}                   &  41.0  \\
                       & \multicolumn{1}{l}{Ours \citep{DBLP:conf/wmt/OttEGA18}} &  43.2 \\ \midrule
\multirow{2}{*}{EN$\rightarrow$CS} & \citet{luong-manning-2016-achieving} &  20.7   \\
                       & \multicolumn{1}{l}{Ours \citep{transformer} } &  25.2  \\ \bottomrule
\end{tabular}
\caption{BLEU scores of translation models for constructing the pseudo data.}
\label{table:translation}
\end{table}

\end{document}